\documentclass{article}

\usepackage[utf8]{inputenc}
\usepackage{amsmath, bm}
\usepackage{amssymb,amsfonts}
\usepackage{hyperref}
\usepackage{graphicx}
\usepackage{grffile}

\begin{document}

\title{Time series analysis with dynamic law exploration}
\author{A. Jakovac}
\date{
\emph{\normalsize Wigner Research Centre for Physics, 1121 Budapest, Hungary}\\[0.5em]
\today} %November 16, 2020

\maketitle

\begin{abstract}
  In this paper we examine, how the dynamic laws governing the time evolution of a time series can be identified. We give a finite difference equation as well as a differential equation representation for that. We also study, how the required symmetries, like time reversal can be imposed on the laws. We study the compression performance of linear laws on sound data.
\end{abstract}

\section{Introduction}

Intelligence is hard to define, and a lot of effort is made to bring up a sensible interpretation \cite{FChollet,HGeffner}. The significance of the definition itself is that it provides a framework of thinking, and as such it initiates the possible questions and thus the possible answers, too. 

As an example, in the field of artificial intelligence (AI) the study of the visual cortex of the brain motivated the creation of feed forward (deep) neural networks, in particular the convolutional networks \cite{LeCunet}. Complemented by the mathematical notions like Bayesian analysis lead to numerous brilliant achievements \cite{Schmidthuber}. In all types of tasks, like classification, regression, data compression and generation the present day neural networks have achieved considerable success.

Nevertheless, AI systems face serious challenges. They are still vulnerable to unexpected errors, adversarial attacks \cite{adversarial}, catastrophic forgetting \cite{Kirckpatrick}. Usually long range correlations are hard to maintain with current AI methods (see for example \cite{texture}). All these difficulties make it necessary to rethink the basic concepts of AI over and over again.

As classification is concerned, the basic view of the intelligent actor is that as output it provides the (conditional) probability $p(y|x)$ that a certain class $y$ can be observed in the input $x$ \cite{GBC}. One can give arguments why this approach is correct \cite{GBC}. On the other hand human perception recognize classes not based on a single probability, but recognizing numerous features in the input. If all, or at least most of the features characteristic to a given class can be found in the input, then we confidentally can single out that class as a result of classification. If characteristic features are missing, or, even worse, we find features that contradict to the given class, then we exclude the input from that class. It is not a linear process, but a hard veto.

In \cite{UU} we tried to lay the mathematical foundation of a theory that supports the intelligent perception through features. The goal of the present paper is to apply these ideas for the analysis of time series with linear features. Linear mappings have a long history, from support vector machines \cite{SVM} to Linear Predictive Coding \cite{LPC}. The presented method of identifying laws are close to these techniques, but uses a different line of thought.

This paper is organized as follows. First we overview the relevant and irrelevant features of \cite{UU} in Section \ref{sec:relevant}. Then we recap the treatment of time series with embedding \ref{sec:timeseries}. In the next Section (Section \ref{sec:linmap}) we examine the properties of linear features. Then we apply our technique for musical data compression example in Section \ref{sec:compression}. Finally we close the discussion with a Conclusions section (Section \ref{sec:conclusions}).

\section{Relevant and irrelevant features}
\label{sec:relevant}

Intelligence can be thought to be an ability to rephrase a complex input in terms of  new notions that fit best to the input set. For example the notion of quadrupedality applies for most mammals, the carnassial dentition is characteristic for carnivora, while the property to be rided applies mostly for horses (or, eventually donkeys, ostriches, water buffalo etc.). But a useful propery is also that a given person has a name Betty; although there are a lot of Betties in the world, but in our closer environment probably there are just few.

Thus a property is useful if either it characterizes a whole set, taking a constant value on each of its elements (for example the property of having caranssial dentition is ``true'' for carnivores and ``false'' for other animals); or if it can be used to make distinction between elements within the set (she is the only person among our friends called Betty). These useful properties can be called \emph{features}, and, following \cite{UU}, they are called relevant, if they are constant on a set, and irrelevant if they can tell elements of a set apart.

For classification the relevant features are the most useful, since each relevant feature singles out a set, the common appearance of several relevant features single out the intersection of the given sets, thus restricting the interpretation of the input. On the other hand for compression (when we know the set) we need to remember only the irrelevant features; for example we can tell our friends to call Betty without needing to specify other properties of the person.

The mathematical basis to deal with relevant and irrelevant features are published in Ref\@. \cite{UU}, here we only recall the most important findings. In this approach the probability space of the intelligent actor is $(X, F_X, \mathcal{P})$, where $X$ is the set of all possible inputs (considered to be a finite set), $F_X$ is its power set, and $\mathcal{P}$ is the uniform discrete probability measure (i.e. $\mathcal{P}(A\subset X) = |A|/|X|$). The intelligent actor (e.g. a neural network) is a set of random variables $\{\xi_i|i=1,\dots,N\}$ over this probability space\footnote{Thus the output is \emph{not} the probability distribution, as it is usual in a lot of works.}, with joint probability distribution over $p^{(A)}(\xi_1, \dots, \xi_N)$ over $A$. To represent a subset $A\subset X$ we choose random variables that are independent over $A$, i.e. the joint probability distribution factorizes $p^{(A)}(\xi_1, \dots, \xi_N) = p_1^{(A)}(\xi_1)\dots p_N^{(A)}(\xi_N)$. Moreover these random variables should be either constant $\xi_i(x\in A)=\sigma_i$, or they have uniform distribution (i.e. the distribution is constant $p^{(A)}(\xi_i=\sigma)=1/|A|$). The former are called \emph{relevant feature}, the latter \emph{irrelevant feature}. One can prove that any set can be completely described by giving appropriate number of relevant and irrelevant features \cite{UU}.

In practice instead of determining the exact relevant features, we are usually satisfied with an approximation. This means that instead $\xi_i(x\in A)=\sigma_i$ constant we can only ensure $|\xi_i(x\in A) -\sigma_i|<\varepsilon$ for some small $\varepsilon$. 

A relevant feature restricts the possible input set to be in $A\subset \xi_i^{-1}(\sigma_i)$, where $\xi_i^{-1}$ is the inverse image. In the approximate case we have $A\subset \xi_i^{-1}([\sigma_i-\varepsilon, \sigma_i+\varepsilon])$. Several relevant features restrict more and more the possible elements of the input set, and so it is the basic building block of classification.

Irrelevant features, on the other hand, are useful to tell apart the elements of the studied subset of input. Indeed, if we denote the range $\xi_i(A)$ by $R$, then $\xi_i^{-1}(r),\ r \in R$ yields a pairwise disjoint subdivision of our subset. Using independent irrelevant features, these subdivisions will differ, and so with enough number of irrelevant features all elements of $A$ can be uniquely described. Thus the task of compression is equivalent to find all the irrelevant features.

Following this line of thoughts we can think about an intelligent actor as one that is capable to find the best features of the studied input set. Depending on the task one should seek the relevant or the irrelevant ones.

%In physics one can consider the inputs as results of observations or measurements. Then a relevant feature, yielding zero for all observations $\mathcal{F}(x)=0$ is called a \emph{physical law}. Most of the laws, however, are not exact, at least not in the environment where we observe them. The difference is usually associated with noise, but, in fact, it is a (small) failure of the law. We see that there is a one-to-one correspondence between the (approximate) physical laws and the (approximate) relevant features.

In the following we discuss the application of the above ideas for the treatment of time series. We will mostly concentrate on relevant features, or laws, because it leads to a recursive description of the series, from where one can restore the complete series. 

\section{Time series, laws and symmetries}
\label{sec:timeseries}

Time series are mathematically $y:\mathbb{R}\to\mathbb{V}$ functions, where $\mathbb{V}$ is a finite dimensional Hilbert space. To be able to handle this function, we have to prepare a finite, representative set from it. We will choose a finite set $\{t_k\,|\, k \in\{1,\dots,K\}\}\subset \mathbb R$ as initial points, and then we will sample the time series at finite past points starting from $t_k$. In this way we generate the database
\begin{equation}
  \label{Y0def}
  {\cal Y} = \{\, Y^{(k)} \in \mathbb V^{n+1} \;|\; Y^{(k)}_i = y(t_k - i\Delta t),\; i\in \{0,\dots,n\},\; k\in\{1,\dots, K\}\,\}.
\end{equation}

We will need to find relevant features (laws), which satisfy
\begin{equation}
\mathcal{F}:\mathbb{V}^{n+1}\to\mathbb{R},\qquad  \mathcal{F}(Y^{(k)})=0,\quad\forall k.
\end{equation}

\begin{figure}[htbp]
  \centering
  \includegraphics[height=4cm]{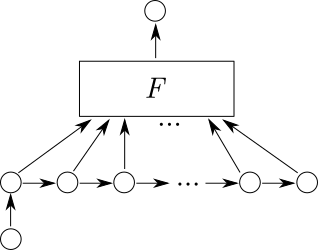}
  \caption[neural network for recursion]{Neural network representation of the recursion. The nonlinear $\mathcal{F}$ function can be itself a neural network.}
  \label{fig:recursion_network}
\end{figure}

This construct can be thought of as a neural network (c.f. Fig. \ref{fig:recursion_network}), if we choose $t_k=k \Delta t$.
In this case we can build a network, where we have $n$ memory slots for the past $i=1,\dots n$ elements, and we always put the new element into the first (0th) place, shifting all the others down. The content of the $i$th slot after $k\ge n$ steps is $Y^{(k)}_{i}$, and the single output is simply
\begin{equation}
  \mathrm{output}^{(k)}= \mathcal{F}(Y^{(k)}_0,\dots,Y^{(k)}_{n}).
\end{equation}

We may have restrictions on the actual form of $\mathcal{F}$ from requirements (prior knowledge) about the time series. For example we may want that the law is time reversal symmetric: this may come from the knowledge that basic physical laws are time reversal invariant, and if we want to describe only a steady state then the state also does not single out an arrow of time. Time reversal symmetry requires that if we feed the function in a reverse direction, the same law applies:
\begin{equation}
  \label{eq:time_reversal_symmetry}
  \mathrm{\mbox{time reversal symmetry:}}\qquad \mathcal{F}(x_0,\dots,x_{n-1}) = \mathcal{F}(x_{n-1},\dots,x_0).
\end{equation}

In some cases, like acoustic signals, or market prices we also expect that the law is independent of the amplitude. Therefore if $x$ satisfies the law, then $\alpha x$ should also satisfy it:
\begin{equation}
  \label{eq:scaling_symmetry}
  \mathrm{\mbox{scaling symmetry:}}\qquad \mathcal{F}(x) = 0 \quad\Rightarrow\quad \mathcal{F}(\alpha x)=0.
\end{equation}
This requirement can be fulfilled if
\begin{equation}
  \mathcal{F}(\alpha x) = \alpha^z \mathcal{F}(x),
\end{equation}
i.e. $\mathcal{F}$ is a scaling function with (an arbitrary) scaling dimension $z$. This requires that all activation functions are power functions.

\section{Linear mapping}
\label{sec:linmap}

The simplest form of a law is a linear map, then the network is similar to the linear support vector machine, albeit we apply a different logics here.

If the maps of the feature function are linear, then it is worth to apply the linear algebra notations. In particular we denote the input set as a matrix
\begin{equation}
\label{Ydef}
Y_{ki} = Y^{(k)}_i = y(t_k - i\Delta t),\qquad k\in\{0,\dots,K\}, \; i\in \{0,\dots,n\}.
\end{equation}
A linear map is equivalent to giving a vector $w\in \mathbb{V}^{n+1}$ for which
\begin{equation}
	\mathcal{F}(z\in\mathbb{V}^{n+1}) = \sum_{i=0}^{n} z_i\cdot w_i \equiv z^T w,
\end{equation}
where the first dot denotes the $\mathbb{V}$ scalar product, while in the second form we use the usual matrix notation. Applying it to our sample set we get $K$ results
\begin{equation}
	\mathcal{F}(Y^{(k)}) = \sum_{i=0}^{n} Y_{ki} w_i \equiv (Yw)_k
\end{equation}
by denoting it as a matrix multiplication.

We remark here that we could use a more general Ansatz here. If we define a coefficient matrix $R\in \mathbb{V}^{n+1}\otimes \mathbb{V}^{n+1}$, then the form of the law could be
\begin{equation}
  \mathcal{F}(z) = z^T R w.
\end{equation}
If $R$ is invertible, then this simply redefines the $w$ weights. Practical use of this form comes when $R$ is diagonal and have zero and one elements.
\begin{equation}
  \mathcal{F}(z) = \sum_{i=0}^{n} r_i z_i\cdot w_i,\qquad r_i \in \{0,1\}.
\end{equation}
In this way we can leave out certain past points from the consideration, which can be a useful technique to reduce the number of effective weights. Technically this can be taken into account with $Y_{ki}\to r_i Y_{ki}$, and in the resulting weights we put $w_i\to r_i w_i$. Thus this general case can be completely embedded in the present discussion.

The above map is a relevant feature/law if $\mathcal{F}(Y^{(k)})=0$ for all $k\in\{0,\dots,K\}$, it is
\begin{equation}
\label{Ylaw}
Yw =0.
\end{equation}
This is equivalent to $|Yw|=0$ with some norm. With the usual quadratic norm we may define
\begin{equation}
	\chi^2 = \frac1K (Yw)^T(Yw) = w^T C_{PCA} w \stackrel!= 0,
\end{equation}
where
\begin{equation}
  C_{PCA} = \frac1K Y^TY
\end{equation}
is the correlation matrix of the given problem (the same matrix appears in the Principal Component Analysis, PCA).

This can be simply satisfied with $w=0$. To avoid this trivial solution we require $|w|=1$. To implement this constraint we realize that the minimum of $\chi^2$ is zero, so instead of an equation, we can think the problem as minimization. Using the Lagrange multiplicator method we should minimize
\begin{equation}
\chi^2_\lambda = w^T C_{PCA}w -\lambda w^Tw = \mathrm{minimal.}
\end{equation}
The solution of this requirement is
\begin{equation}
  C_{PCA} w^{(\lambda)} = \lambda w^{(\lambda)}
\end{equation}
eigenvalue equation. Since $Y^TY$ is symmetric and positive definite, all eigenvalues are real and positive. Rewriting it to $\chi^2$ we find
\begin{equation}
\chi^2 = \lambda.
\end{equation}
This means that an exact law would be equivalent to find a zero eigenvalue in $C_{PCA}$.

In practice it is usually not possible to satisfy, we always have a (small) breaking of the law. We can describe it as
\begin{equation}
  \mathcal{F}(Y^{(k)})= \xi_k,
\end{equation}
where $\xi_k$ can be thought of as a random process. It is remarkable that
\begin{equation}
  \left\langle \xi^2 \right\rangle = \frac1K \sum_{k=1}^K \xi_k^2 = \chi^2 =\lambda.
\end{equation}
This means that the variance of the accuracy of the law is given by the eigenvalue.

We remark that although in the PCA analysis we use the same matrix, but there we seek the largest eigenvalue. The PCA logic is to find the direction that aligns with the data the most, i.e. the direction where the data vary the most, which corresonds to the largest eigenvalue of the PCA matrix. We, however, look for a direction that characterizes all the data, so we need the direction where the data differ the least. This corresponds to the smallest eigenvalue of the PCA matrix. In the language of \cite{UU}, the eigenvectors belonging to the largest eigenvalues are the irrelevant, the ones belonging to the smallest eigenvalues are the relevant features.

\begin{figure}[htbp]
  \label{fig:PCA_spectrum}
  \begin{center}
    \includegraphics[height = 5cm]{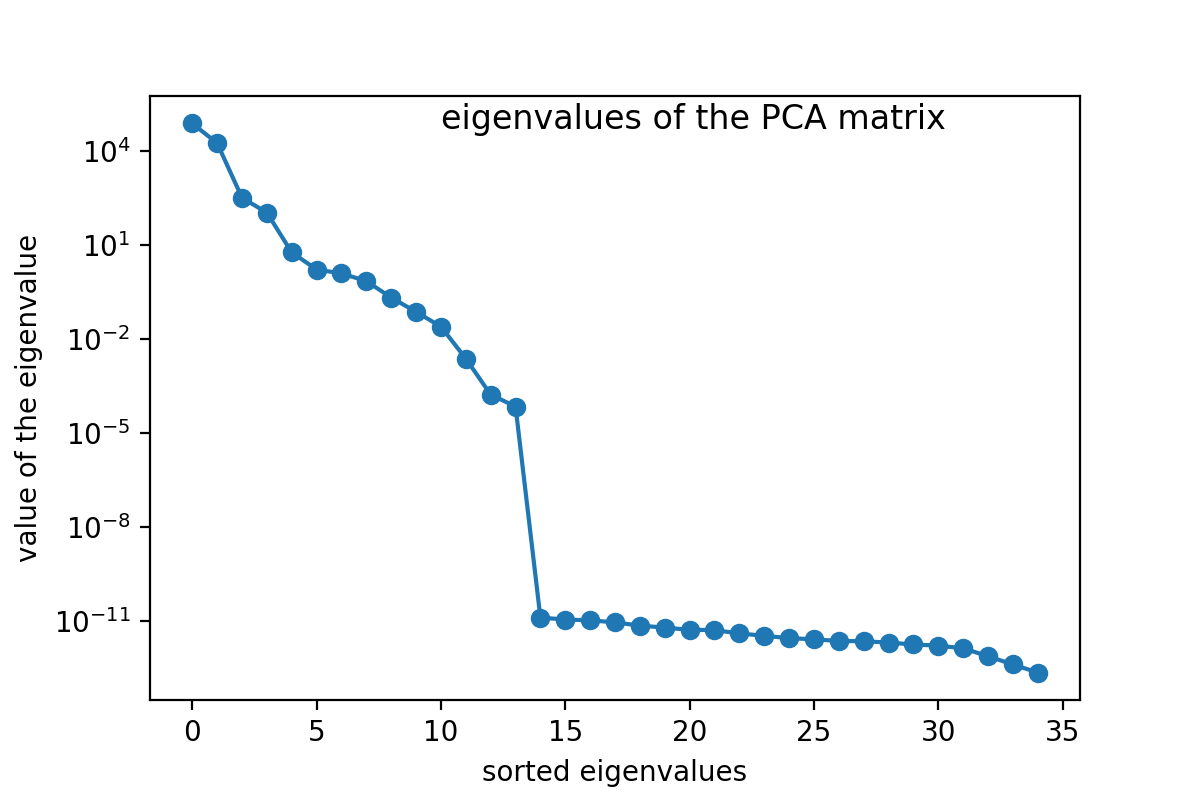} 
  \end{center}
  \caption{Spectrum of the PCA matrix for $N_{law}=35$ for the sound of the flute. Small eigenvalues correspond to laws, i.e. general restrictions on the data.}   
\end{figure}

In realistic situations when we analyze the eigensystem of the PCA matrix we find a spectrum, similar to what can be seen in Fig. \ref{fig:PCA_spectrum}.
As we see, the PCA matrix has a few large(ish) eigenvalues and a lot of small ones. If we went on to determine longer laws, we would have determined more and more small, i.e. relevant eigenvalues. This is a very general observation for natural processes: we have much more relevant than irrelevant features \cite{UU}. Any of the relevant features can be used as a law.

\subsection{Function class compatible with linear law}

A generic matrix does not have a zero eigenvalue. Let us examine what properties a time dependent function must have to make it possible. Here we restrict ourselves to $\mathbb{V} = \mathbb{R}$ real valued time series.

A law over real valued time series can be organized as a recursion. First, using the definition of $Y$ \eqref{Ydef} we can rewrite the law \eqref{Ylaw} as
\begin{equation}
y(t_k) = -\frac1{w_0} \sum_{i=1}^{n} y(t_k - i\Delta t) w_i.
\end{equation}
Now choosing $t_k=k\Delta t$ we can define a series
\begin{equation}
y_k = y(k \Delta t).
\end{equation}
It satisfies then
\begin{equation}
\label{recursion}
y_{k} = -\frac1{w_0} \sum_{i=1}^{n} y_{k-i} w_i,\qquad \forall k\ge n.
\end{equation}

Being a linear recursion, sum of solutions is also a solution. For a particular solution we try an Ansatz
\begin{equation}
y_k = q^{-k}.
\end{equation}
Rewriting this Ansatz into \eqref{recursion} we find
\begin{equation}
0 = \sum_{i=0}^{n} w_i q^{-k+i} = q^{-k} \sum_{i=0}^{n} w_i q^{i}.
\end{equation}
This is true, if
\begin{equation}
\sum_{i=0}^{n} w_i q^i = 0.
\end{equation}
That means that the solutions are roots of the 
\begin{equation}
\label{polynomial}
\mathcal{P}(x) = \sum_{i=0}^{n} w_i x^{i} = 0
\end{equation}
$n$th order polynomial. This equation yields $n$ solutions, that are either real, or complex, appearing in conjugated pairs (so altogether $n$ real data). Let us denote the roots as $\{q_a\,|\, a=1\dots n\}$. With all these roots we can fully fit the $n$ initial conditions of the recursion, thus it yields the complete solution.

As a side remark we can establish a relation between $w$s and $q$s from the condition that
\begin{equation}
\label{polynomial_connection}
	\sum_{i=0}^{n} w_i x^{i} = w_n \prod_{a=1}^{n}(x-q_a).
\end{equation}

Thus the time series with generic initial conditions that satisfies the linear law can be written as
\begin{equation}
y(k\Delta t) = \sum_{a=1}^{n} c_a q_a^{-k},
\end{equation}
where the $c_a$ coefficients are complex numbers. With the notation 
\begin{equation}
\label{alphadef}
\alpha_a = \dfrac{1}{\Delta t} \ln q_a
\end{equation}
we can write
\begin{equation}
\label{ytform}
y(t) = \sum_{a=1}^{n} c_a e^{-\alpha_a t}.
\end{equation}
Although originally it is valid for discrete times, but it is easy to verify that the above form is a solution for any starting time. This form gives the definition of that function class that is compatible with a linear law.

To give still another view, we observe that
\begin{equation}
(\partial_t+\alpha) e^{-\alpha t} = 0.
\end{equation}
Therefore \eqref{ytform} satisfies
\begin{equation}
\prod_{a=1}^{n}(\partial_t+\alpha_a) y(t) = 0.
\end{equation}
Thus $y(t)$ satisfies an $n$th order linear differential equations with constant coefficients.

A corollary is that in the linear case a law can be given in multiple ways:
\begin{itemize}
\item give an $w\in\mathbb{R}^{n+1}$ weight function, then the element satisfy the law lie in the orthogonal subspace: $y^Tw=0$
\item give a linear $n$th order recursion: $y_0 = -\sum_{i=1}^n \bar w_i y_i$, where $\bar w_i= w_i/w_0$
\item give the $q_a$ roots of the polynomial $\mathcal{P}(x)= \sum_{i=1}^n w_i x^i$: the weights can be reconstructed from $\sum_{i=1}^n w_i x^i = w_n\prod_{a=1}^n(x-q_a)$
\item give the corresponding differential equation.
\end{itemize}

The recursive form of the law \eqref{recursion} can be generalized to nonlinear laws, then the law should be formulated as a (nonlinear) recursion.

\subsection{Stability of a recursion}

Let us consider the asymptotic behavior of the recursion, i.e. for large $t$. If the real part of $\Re\alpha > 0$, then for large $t$ it yields an exponentially growing solution, that is to be excluded. If $\Re\alpha<0$, then for long times it is vanishing, and we will not observe it. Therefore for long times only the imaginary $\alpha$ survive.

Since $y(t)$ must be real for all $t$, the roots $e^{i\omega t}$ and $e^{-i\omega t}$ both appear in the series (with complex conjugate coefficients). This means that in the differential equation form we find
\begin{equation}
(\partial_t + i\omega)(\partial_t - i\omega) = \partial_t^2+\omega^2.
\end{equation}
Therefore the differential equation is \emph{time reversal invariant}.

What does it imply for the weights in the discretized law? We should consider the relation \eqref{polynomial_connection} which connects the roots with the coefficients. Now the roots appear in pairs:
\begin{equation}
(x-e^{i\omega \Delta t})(x-e^{-i\omega \Delta t}) = x^2 - 2x\cos(\omega\Delta t) +1.
\end{equation}
Here the coefficient of the $x^2$ and $1$ terms are the same. This will be inherited to all of the weights. To see it, we observe that the polynomial coming from the above pairs satisfies (using $Z_a=-2\cos(\omega\Delta t)$)
\begin{equation}
\mathcal{P}(\dfrac{1}{x}) = \prod_{a=1}^\ell (x^{-2} + Z_ax^{-1} +1) = x^{-2\ell} \prod_{a=1}^\ell (1 + Z_ax  +x^2) =  x^{-2\ell}\mathcal{P}(x).
\end{equation}
where $\ell$ is the number of frequencies (i.e. $n=2\ell$). If we consider the same polynomial in weight representation, we can write (note that $2\ell = n$)
\begin{equation}
x^{n}\mathcal{P}(\frac1x) = x^{n}\sum_{i=0}^n w_i x^{-i} = \sum_{i=0}^n w_i x^{n-i} = \sum_{j=0}^n w_{n-j} x^{j}. 
\end{equation}
This is equal to $\mathcal{P}(x)$, if
\begin{equation}
w_{n-i} = w_i,\qquad \forall i=0,\dots n.
\end{equation}
This also means that the recursion forward and backward happens with the same law: this is the manifestation of time reversal symmetry.

This is the property that can be easily generalized to nonlinear recursions: we should require that the forward and backward iteration happens with the same law.

\subsubsection{Example: a single sin}

To see an example, consider a simple function:
\begin{equation}
y(t) = A \sin(\omega t +\varphi).
\end{equation}
This satisfies $(\partial_t^2 +\omega^2)y(t)=0$, so we know that it can be represented a law consisting of 3 terms (it can be longer, too, see later).

Indeed, from the harmonic function addition law
\begin{equation}
y_c = y(t+c\Delta t) =  A \sin(\omega t +\varphi)\cos(c \omega \Delta t) + A \cos(\omega t +\varphi)\sin(c \omega \Delta t)
\end{equation}
we see that
\begin{equation}
y_{-1} -2\cos(\omega \Delta t) y_0 + y_1 = 0.
\end{equation}
Thus the weights are $w=(1, Z, 1)$, where $Z=-2\cos\omega\Delta t$, as it was claimed before.

We could use the relation between the root and weight representation of a polynomial \eqref{polynomial_connection} to find the same result.

\subsubsection{Example: sum of two sin functions}

Let us consider the function
\begin{equation}
y(t) = A_1 \sin(\omega_1 t +\varphi_1)+ A_2 \sin(\omega_2 t +\varphi_2).
\end{equation}
This satisfies
\begin{equation}
  (\partial_t^2+\omega_1^2)(\partial_t^2+\omega_2^2) y = 0.
\end{equation}
The corresponding polynomial reads
\begin{eqnarray}
\sum_{k=0}^4 w_i x^{i} && = (x^2+Z_1x+1)(x^2+Z_2x+1) =\nonumber\\ && =x^4 + (Z_1+Z_2)x^3 + (2+ Z_1Z_2) x^2 + (Z_1+Z_2)x +1.
\end{eqnarray}
We see that the composite law maintains time reversal symmetry, as it should.

\section{Data compression using linear laws}
\label{sec:compression}

Laws, as we have seen, represent a function class, and so they can be used to represent data with appropriate coefficients. Linear laws include polynomials, exponentials and harmonic functions (and also their mixture or product of them), and thus they can replace existing methods with broader possibilities.

If the data support our function class, then the law represents it for a long time, in extreme cases all points. We remark that these include also $ax^n e^{-bx}$ like transient or saturation processes besides other examples.

For the description an actual data series which (approximately) satisfy a law, we need also the initial conditions for the recursion. Given the law, i.e. the function class, then each initial condition represents a different element of this class. These are useful to tell apart the elements of the class, and so they correspond to the irrelevant quantities in the language of \cite{UU}.

\subsection{Coefficient fitting}

We can determine the initial conditions in several ways. The first is to determine the $q_a$ roots of the corresponding polynomial. Then we know the the data can be described by the function \eqref{ytform}, we just have to fit the coefficients. In order to do that we introduce the matrix
\begin{equation}
  Q_{ki} = q_i^k,
\end{equation}
and use least square method
\begin{equation}
  \chi^2 = \sum_k (y_k - \sum_i c_i q_i^k)^2 = y^Ty -2 y^T Q c +c^TQ^TQc = \mathrm{minimal.}
\end{equation}
This means that the coefficient vector comes from the solution of
\begin{equation}
  Q^TQ c = Q^Ty.
\end{equation}
Once we have the coefficients, we can propose a continuous function that approximates data the best.

Another method is that we use the recursion directly to describe the data. The problem is that if our law is not exact, i.e. the data are noisy, then we can not just take simply the first $n$ terms to start the recursion. We should start from an optimal $\bar y_k$ starting values in order to achieve the best overall fit to the data.

The best way to find these optimal values is to define a series basis starting from initial condition $y_k = \delta_{k\ell}$ with given $\ell$, for $\ell=0,\dots, n-1$. Using the recursion \eqref{recursion} we will have a series $s^{(\ell)}$. Then we require
\begin{equation}
   \chi^2 = \sum_k (y_k - \sum_{\ell=0}^{n-1} \bar y_\ell s^{(\ell)}_k)^2 = \mathrm{minimal.}
\end{equation}
Introducing a matrix again
\begin{equation}
  S_{k\ell} = s^{(\ell)}_k,
\end{equation}
the initial condition vector $\bar y$ comes from the solution of
\begin{equation}
  S^TS \bar y = S^T y.
\end{equation}

In case of nonlinear laws we have to use the recursive form of the law to determine the best description of the data. There the sum of solutions is not a solution and so the initial conditions should be fitted.

\subsection{Results}

Usually the accuracy of a single law will decrease with increasing data length. An $n+1$ long linear law means $n$ roots and $n$ amplitudes, together $2n$ parameters: it can therefore describe a $N_{data}=2n$ long data vector. For longer vectors we unavoidably loose precision, the question is only, what is the rate of this loss. To numerically characterize the loss, we can plot the accuracy $A$ versus the compression rate $R$, defined as
\begin{equation}
  A = 1-\frac{|X-X_{sym}|}{|X|}, \qquad R =\frac{N_{data}}{2n},
\end{equation}
where $X$ is the data which is $N_{data}$ long, $X_{sym}$ is the simulated data using the $n+1$ long law. We always get $A=1$ for $R=1$.

\begin{figure}
  \label{fig:random}
  \begin{center}
    \includegraphics[height = 5cm]{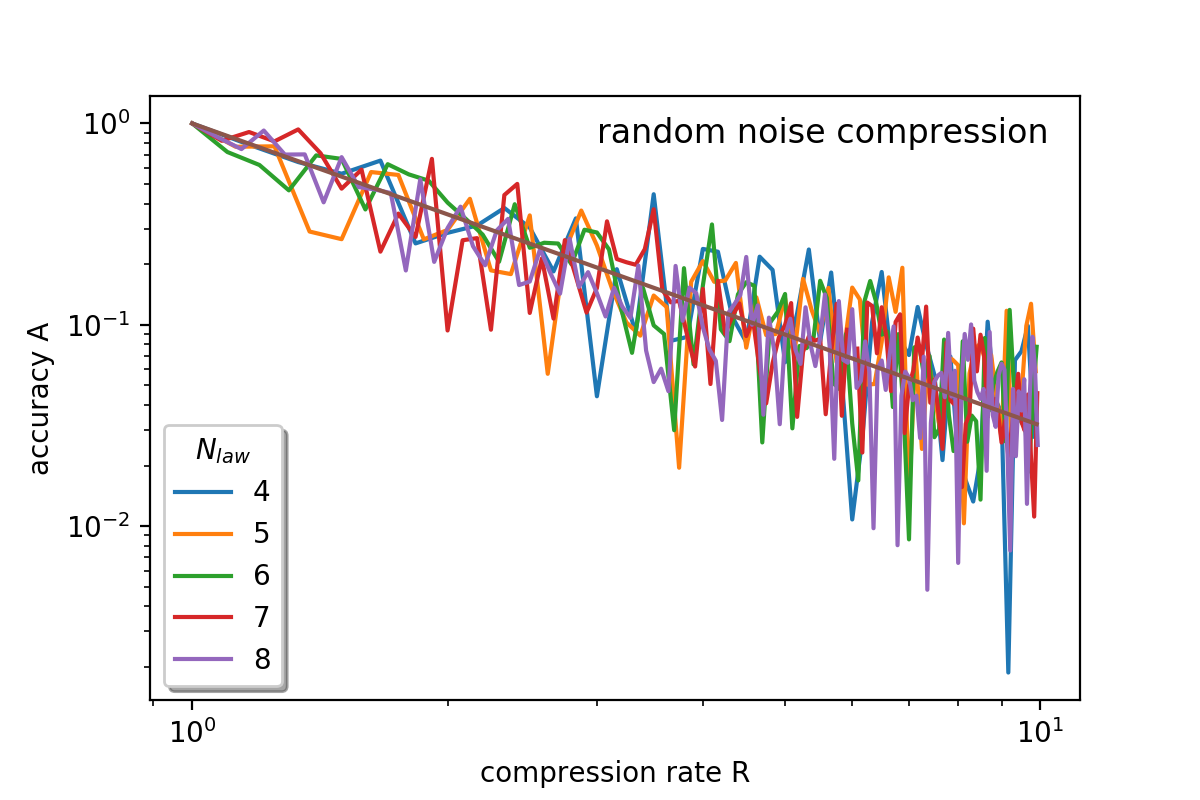}
  \end{center}
  \caption{Accuracy vs. compression rate for random noise, in log-log plot. The straight line is the $A=R^{-3/2}$ relation.}
\end{figure}

An extreme situation is when the data exactly support our function class: there $A=1$ for any $R$. The other extreme is the random data series, where there is no law at all, see Fig. \ref{fig:random}. There we can observe that $A = R^{-3/2}$ is the average trend, independently on the number of laws we use.

\begin{figure}
  \label{fig:flute}
  \begin{center}
    \hspace*{-1em}\hbox{
      \includegraphics[height = 4.2cm]{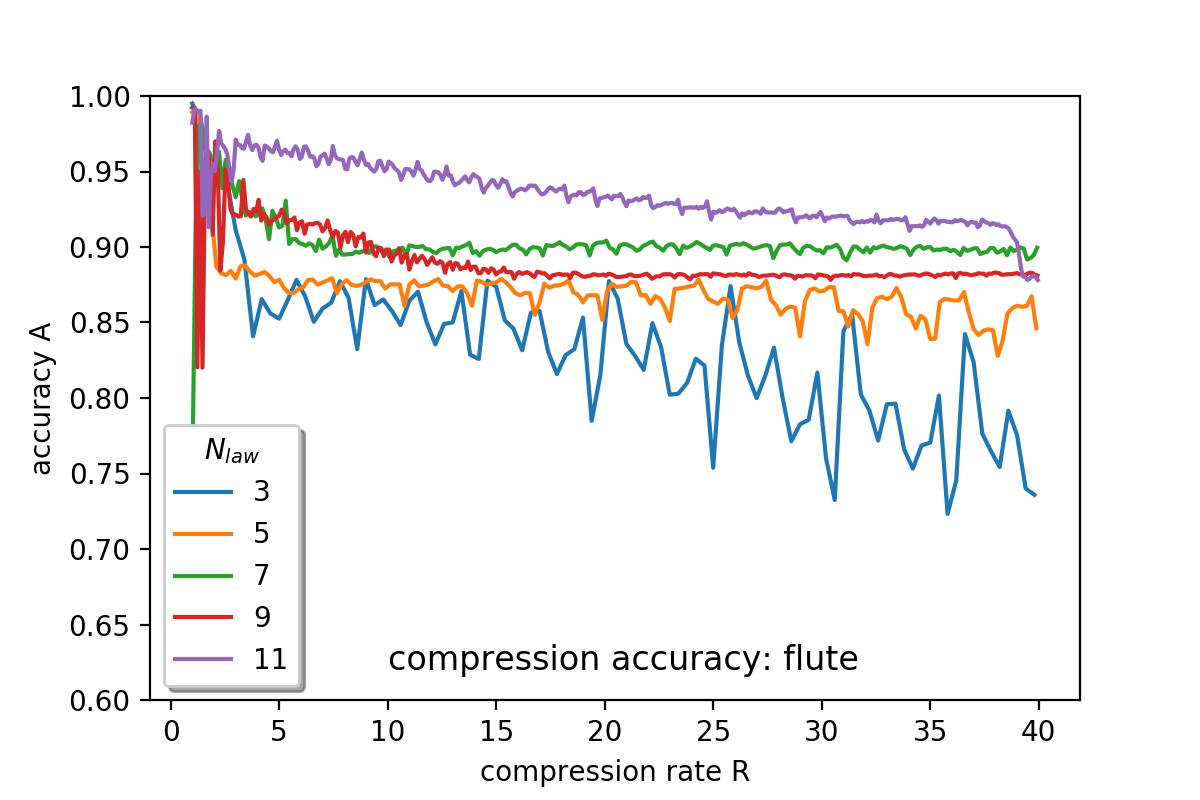}
      \includegraphics[height = 4.2cm]{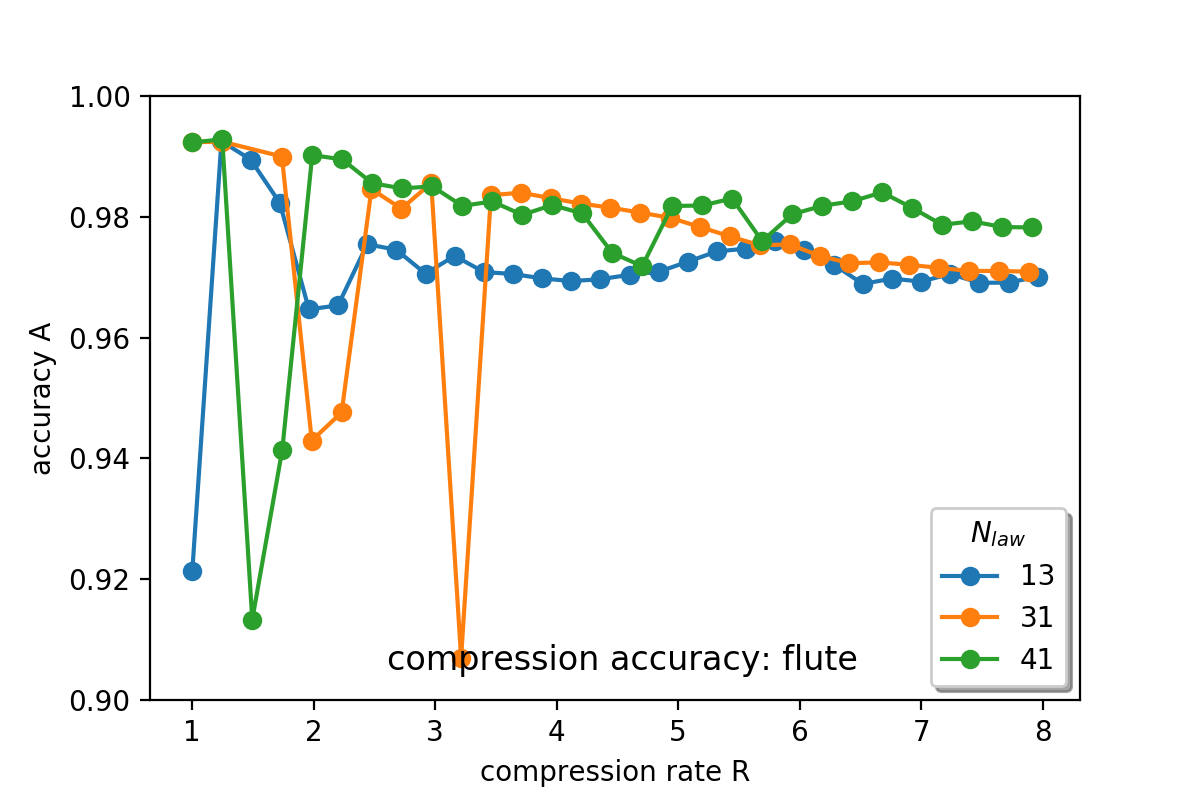}
    }
  \end{center}
  \caption{Accuracy at given compression rate for flute. In the left and right panel we used different compression rate range and different $N_{law}$. In both panel $\Delta t= 4/$sampling rate.}
\end{figure}

In all intermediate situation we lose precision, the rate depends on the actual data. As an example we study the sound of a flute, and determine the accuracy of the representation. It may remain close to a perfect desciption for a large compression rate, as we can see from Fig. \ref{fig:flute}.
As it can be seen, even for very large compression rates of 40, the best laws may provide reasonable 90\% accuracy. It is not the level of HiFi acoustic experience, but for recognizing a sound it is perfectly usable. 

\begin{figure}
  \label{fig:best}
  \begin{center}
    \includegraphics[height=5cm]{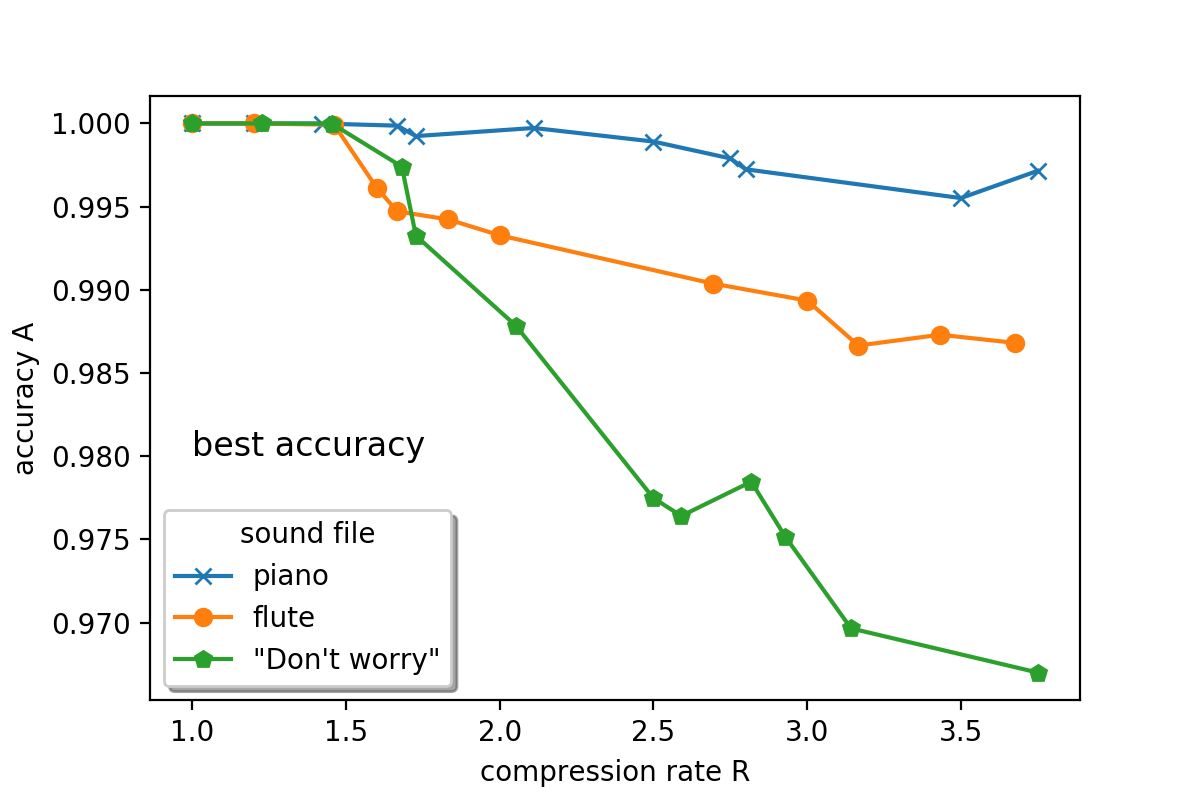}
  \end{center}
  \caption{The available best accuracy for different music, coming from optimizing $N_{law}$ and $\Delta t$.}
\end{figure}

Usually a the best accuracy for a given compression rate comes from an appropriate choice of the number of terms in the law and $\Delta t$. In case of clear acoustic data, like the sound of the flute, one can reach a large compression, while for more noisy situation, like in percussion instruments, it is hard to increase the compression rate too much, c.f. Fig. \ref{fig:best}.
The problem with percussion instruments (even if the are imitated by human voice, like in Bobby McFerrin's "Don't worry" song) is that they consist mainly of "switch-on" effects, when the device starts to emit sound. These are sudden and short burst, which means that the laws change frequently. The linear law method we can only give a homogeneous law, so it will have difficulty to represent these musical instruments. Nevertheless, the nearly 97\% accuracy in the randomly chosen part of the song, at compression rate about 4, is again not a bad description. It can also be noted that for small compression rate (up to about 1.5), the compression is parctically lossless. This property can be used to lossless compression techniques.

\subsection{Comparison with Linear Predictive Coding technique}

Linear Predictive Coding (LPC), applied in many areas of sound file compression (e.g. FLAC \cite{FLAC}), uses a similar formalism. There are, however some important differences. LPC uses a recursive form with $w_0=-1$ prescription for the weight, which also means that the $w_{n>0}$ coefficients must be determined by matrix inversion. In our case the weights are eigenvectors belonging to the smallest eigenvalue, which is more stable procedure. A further difference is that we only determine the dominant frequencies (relevant features), the actual coefficients of the recursion comes with an extra coefficient fitting step.

Another difference is that LPC wants to predict the next data point, while we give a continuous model that describes it. As a consequence we are not forced to use $\Delta t=1/$sampling rate, which extends our possibilities. Usually the best compression comes from choosing not the smallest $\Delta t$.

In addition, a law can be formed to satisfy symmetry requirements, like time translational symmetry.

\section{Conclusions}
\label{sec:conclusions}

The goal of this paper was to demonstrate how the idea of finding laws in input data sets is working in the linear case. We used time series to analyze, in particular sound tracks, embedded in finite $n+1$ dimensions.

Linear laws can be represented by an $n+1$ dimensional vector that is perpendicular to all embedded data. These vectors can be found as the eigenvectors belonging to the zero eigenvalue of the corresponding PCA problem. In realistic case the data do not support the law exactly. In this case we choose the smallest eigenvalue, then we know that the variance of the law breaking term is equal to the value of this eigenvalue.

As it was shown, the function class that satisfies a linear law consists of the solutions of $n$th order differential equations with constant coefficients. We determined the relation between the law and the coefficients, which allows us to give a continuous function that describes the data.

Laws can be constrained to satisfy certain requirements. Long time stability, for example, requires time reversal symmetry, which is manifested in the weights as a symmetry under reversing it.

These ideas are checked to describe realistic data, in this paper we considered music instruments. We plotted the accuracy of the description of the data vs. the compression ratio. Depending on the data the efficiency of linear laws are different. In the lack of laws, like in random data set, the accuracy decreases very fast. The more the data support a linear law, the longer a high precision maintains. Among our examples the best compression rate could be achieved in case of piano music.

A lot of finding of the linear laws can be taken over for the nonlinear case. In particular in neural networks the last layer, which is usually a fully connected one, can be substituted by a linear law finder. Then in the result we will find several optimized features, corresponding to the corresponding laws. With more relevant features hopefully the classification tasks can be performed more reliably, since it is more and more difficult to find such adversarial directions that satisfy all the laws, but still takes us out of the given class.

\section*{Acknowledgment}

The author acknowledges important discussions with T.S. Biro, D. Berenyi, P. Mati, P. Posfay and A. Telcs.
The research was supported by the Ministry of Innovation and Technology NRDI Office within the framework of the MILAB Artificial Intelligence National Laboratory Program, and the Hungarian Research Fund NKFIH (OTKA) under contract No. K123815.

\appendix
\section{Relation between root finding of a polynomial and diagonalization}

Consider a recursion
\begin{equation}
y_{N} = \sum_{k=1}^{n} c_k y_{N-k}.
\end{equation}
This recursion can be written two ways.

First we may assume that the solution of the recursion is
\begin{equation}
y_N= q^{-N},
\end{equation}
which yields
\begin{equation}
  \sum_{k=0}^{n} c_k y_{k} =0
\end{equation}
with $c_0=-1$.

Second we can set up a matrix form with
\begin{equation}
Y_N = \left(\begin{array}{l} y_N\cr y_{N-1}\\ \vdots \\ y_2\\ y_1
\end{array}\!\right),\qquad 
M = \left( 
\begin{array}{ccccc}
c_1 & c_2 & \cdots & c_{n-1} & c_{n} \\
1   &  0  & \cdots & 0       &  0 \\
\vdots  &  \vdots  & \ddots & \vdots &  \vdots \\
0   &  0  & \cdots & 0       &  0 \\
0   &  0  & \cdots & 1       &  0 \\
\end{array}
\right)
\end{equation}
as
\begin{equation}
Y_N = M Y_{N-1}.
\end{equation}

The $M$ matrix is diagonalizable $M=U^{-1}DU$, then with $Z=UY$ we find
\begin{equation}
Z_N = D Z_{N-1} = D^N Z_0 = \left( 
\begin{array}{ccccc}
\lambda_1^N & 0 & \cdots & 0 & 0 \\
0   &  \lambda_2^N  & \cdots & 0 & 0 \\
\vdots  &  \vdots  & \ddots & \vdots &  \vdots \\
0   &  0  & \cdots & \lambda_{n-1}^N &  0 \\
0   &  0  & \cdots & 0 &  \lambda_n^N \\
\end{array}
\right)
\left(\begin{array}{l} z_0\cr z_{-1}\\ \vdots \\ z_{-n-2}\\ z_{-n-1}
\end{array}\!\right).
\end{equation}
Thus\begin{equation}
z_{N+1-k} = \lambda_k^N z_{1-k}.
\end{equation}
The original $y_N$ is a linear combination of these expressions, any of this is a solution of the recursion. Therefore the roots are equal to the eigenvalues
\begin{equation}
	q_a = \lambda_a.
\end{equation}

\end{document}